\documentclass{article}

\usepackage{PRIMEarxiv}

\usepackage[utf8]{inputenc} 
\usepackage[T1]{fontenc}    
\usepackage{url}            
\usepackage{booktabs}       
\usepackage{amsfonts}       
\usepackage{nicefrac}       
\usepackage{microtype}      
\usepackage{lipsum}
\usepackage{fancyhdr}       
\usepackage{graphicx}       
\usepackage[hidelinks]{hyperref}
\usepackage{orcidlink}
\graphicspath{{media/}}     

\title{From Pampas to Pixels: Fine-Tuning Diffusion Models for Gaúcho Heritage
}

\author{
  Marcellus Amadeus \orcidlink{0009-0002-7777-2562}\\
  Alana AI Research \\
  São Paulo, Brazil\\
  \texttt{marcellus@alana.ai} \\
   \And
   William Alberto Cruz Castañeda \orcidlink{0000-0002-9803-1387}\\
  Alana AI Research\\
  São Paulo, Brazil\\
  \texttt{william.cruz@alana.ai} \\
    \And
  André Felipe Zanella \orcidlink{0000-0001-8828-6629}\\
  Alana AI Research\\
  São Paulo, Brazil\\
  \texttt{andre.zanella@alana.ai} \\
    \And
  Felipe Rodrigues Perche Mahlow \orcidlink{0000-0001-9816-1440}\\
  Alana AI Research\\
  São Paulo, Brazil\\
  \texttt{felipe.mahlow@alana.ai} \\
}

\begin{document}
\maketitle

\begin{abstract}
Generative AI has become pervasive in society, witnessing significant advancements in various domains. Particularly in the realm of Text-to-Image (TTI) models, Latent Diffusion Models (LDMs), showcase remarkable capabilities in generating visual content based on textual prompts. This paper addresses the potential of LDMs in representing local cultural concepts, historical figures, and endangered species. In this study, we use the cultural heritage of Rio Grande do Sul (RS), Brazil, as an illustrative case. Our objective is to contribute to the broader understanding of how generative models can help to capture and preserve the cultural and historical identity of regions. The paper outlines the methodology, including subject selection, dataset creation, and the fine-tuning process. The results showcase the images generated, alongside the challenges and feasibility of each concept. In conclusion, this work shows the power of these models to represent and preserve unique aspects of diverse regions and communities. 
\end{abstract}


\section{Introduction}
In recent years, the domain of computer science and artificial intelligence (AI) has witnessed significant advancements in the discipline of Generative AI (GenAI). GenAI focuses on creating models capable of generating text, audio, and images based on patterns learned from training datasets. Generative models like GPT-4 \cite{openai2023gpt4} have demonstrated remarkable abilities to produce coherent and informative text, enabling applications in chatbots, machine translation, and the high-quality content creation. In the audio field, notable generative models such as WaveGAN \cite{donahue2019adversarial} have gained prominence. Furthermore, recent advancements in Text-to-Speech (TTS) models have contributed to the evolving landscape of audio synthesis \cite{kaur2023conventional, lobato2023performance}. Within the image generation field, groundbreaking models such as DALL·E 3 \cite{betker2023improving, ramesh2021zeroshot}, MidJourney \cite{midjourney}, and Stable Diffusion \cite{rombach2022high} are capable of generating images from textual descriptions. Applications of these models are found in graphic design \cite{hughes2021generative}, augmented reality \cite{liu2020arshadowgan}, and more, underscoring the significant impact that Generative AI has on creativity and innovation across diverse fields.

Text-to-image (TTI) models refer to a research and development area aimed at creating methods and algorithms capable of converting written text into visual images. This approach seeks to establish a bridge between natural language and visual representations, exploring ways to transform textual descriptions into coherent and meaningful visual content. A pivotal advancement in this domain lies in the emergence and adoption of Latent Diffusion Models (LDMs) \cite{rombach2022high}. These models have become instrumental in crafting images, especially those of artistic and cultural relevance. Rooted in the foundational framework of Diffusion Probabilistic Models (DPMs) \cite{ho2020denoising}, LDMs have wrought a transformative impact on the panorama of image generation, facilitating the generation of high-resolution images while achieving exceptional diversity and photorealism. LDMs have been increasingly utilized in various domains, such as in the medical field for generating brain images \cite{pinaya2022brain}, reconstructing images from brain activity \cite{takagi2023high}, and chest X-rays \cite{packhauser2023generation, weber2023cascaded}. They have also been used for video generation \cite{zhou2022magicvideo, he2022latent}, and more broadly, are being proposed as strategies in architecture \cite{yildirim2022text, ploennigs2023ai}, news illustration \cite{liu2022opal}, and in nursing education \cite{reed2023ai}. 

Among the broadest possible applications of GenAI, we will focus on three main aspects in this work. Firstly, we assess the potential of these models as tools for representing local concepts and communities. Secondly, we will address the representation of historical figures and moments, especially in cases where visual representations are scarce, with historical events and individuals often lacking substantial visual depictions. Thirdly, we will examine the feasibility of representing endangered animals and plants, which can also pose a significant challenge in capturing diverse images. To achieve this, we exemplify the application of TTI in the cultural heritage of Rio Grande do Sul (RS), Brazil. This regional culture, known for its historical, traditional, and visually unique attributes, embodies a rich repository of cultural symbols, costumes, breathtaking landscapes, and pivotal historical figures that significantly shape the cultural identity of southern Brazil. Our work seeks to demonstrate the feasibility of employing diffusion models to generate images that authentically represent the cultural and historical value of Rio Grande do Sul. We will present the image generation process along with the obtained results.

The work is organized as follows: In Section \ref{sec:Method}, we describe the methodology to conduct the study, including the definition of training subjects, datasets, and training process; Section \ref{sec:Exp} summarizes the details of the experimental process carried out in our case study, to validate our proposed methodology. In the Results Section \ref{results} we present the practical application of these techniques in three key areas: Cultural Aspects, Fauna and Flora, and the Farroupilha Revolution, presenting the importance of each concept and the challenges faced for each one. Finally, we conclude the work by presenting our conclusions in Section  \ref{conclusion}, as well as perspectives for future improvements.

\section{Method}
\label{sec:Method}

Recent advances in TTI models allow fine-tuning of pre-trained models to inject a specific subject into the output domain of the model, i.e., the model can generate diverse instances of the same subject. This section describes subject-driven models using existing pre-trained models and open-source libraries.

\subsection{Training topics}

The subject selection is a pivotal step as it determines the model trend toward the type of images representing that subject. Subject-driven generation allows development in different contexts while maintaining the distinctive features of the concept. There are no restrictions on the choice of concepts. They can range from simple objects and living beings (specific individuals or animals) to complex concepts such as war battle paintings. However, they may require additional effort in subsequent steps, including image selection to compose a dataset and fine-tuning experimentation.

\subsection{Dataset creation}

Within the contexts defined, the next step involves selecting relevant images to compose a dataset for each specific scenario. The quantity of samples to cover a dataset is associated with the complexity of the concept. The first concern is to assess the image availability related to the specific subject. If there is a shortage of images, the dataset will restricted to a few available images. A web crawling algorithm can be adopted to automate the task across available image databases. 
Data curatorship after collection and annotation is also desirable, as it involves discarding out-of-context images and also poor resolution and low DPI.

\subsection{Fine-tuning process}

Fine-tuning allows the model to adapt to specific nuances and characteristics associated with individual concepts. The Diffusers \cite{von-platen-etal-2022-diffusers} library offers fine-tuning scripts for training pre-trained diffusion models, enabling subject-driven generation. In addition to standard TTI training, specific techniques designed for this purpose, such as DreamBooth \cite{ruiz2023dreambooth}, are available within the framework.

DreamBooth, introduced in 2022 by a Google Research Team \cite{ruiz2023dreambooth}, represents one of the most widely adopted methods in TTI synthesis. This approach addresses the growing demand for personalized image generation by fine-tuning pre-trained text-to-image diffusion models. By associating unique identifiers with specific subjects, DreamBooth enables the creation of photorealistic images set in various contexts, all guided by user-defined text prompts. Notably, it expands the model's language-vision dictionary to encompass user-specified subjects and leverages autogenous, class-specific prior preservation mechanisms to support the generation of diverse instances within the same class. The applications of DreamBooth extend to subject recontextualization, property modification, and creative art generation. Despite being very recent, DreamBooth has already been used to generate automotive images \cite{sutedy2022text}, to improve the performance of CNNs \cite{zhang2023dreambooth} and 3d generation \cite{raj2023dreambooth3d}. The DreamBooth script is available at Diffusers\footnote{https://github.com/huggingface/diffusers/tree/main/examples/dreambooth} and available to train using any diffusion-based TTI model, such as Stable Diffusion 1.5 or Stable Diffusion XL.

Fine-tuning TTI models are costly and demand powerful GPUs. To mitigate this, several strategies exist to optimize GPU RAM utilization, available at Diffusers \cite{von-platen-etal-2022-diffusers}. These techniques enable training with GPUs with less memory. Some of these optimization strategies are enumerated below:


\begin{itemize}
    \item LoRA (Low-Rank Adaptation): LoRA \cite{hu2021lora} can effectively reduce the memory requirements of deep learning models by approximating the weight matrices with low-rank factors, enabling the training of larger models with limited GPU memory.

    \item Gradient Accumulation: Implement gradient accumulation to split the backpropagation process into smaller batches, which can help reduce GPU memory requirements. This approach allows you to accumulate gradients over multiple forward passes before updating model parameters.

    \item Reduce Batch Size: If feasible, consider reducing the batch size during training. Smaller batch sizes require less GPU memory but may result in longer training times. It's important to find a balance between batch size and training time based on specific requirements.

    \item 8-bit Adam: using a lower bit optimizer, or \textit{fp16} for training can significantly
    reduce memory requirements.

\end{itemize}

\section{Experiments}
\label{sec:Exp}

In this section, we describe our case study of creating subject-driven TTI models to represent the regionalism, culture, and historical value of the state of RS, Brazil.

\subsection{Subjects and Datasets}
\label{datasets}

In this work, we study and evaluate the generative capabilities of Stable Diffusion related to concepts from the following categories: regional biome, historical events and personalities, cultural costumes, and traditional attire. As for the biome, the selected concepts include the ``Araucária", a distinctive tree native to the southern region, as well as the ``Gato-do-mato-pequeno" and ``Sapinho-admirável-de-barriga-vermelha", which are characteristic animals of the area. The Farroupilha Revolution is a significant historical event for RS that influenced Brazilian history. We chose to represent historical figures of this conflict: Giuseppe Garibaldi, Anita Garibaldi, and Bento Gonçalves, as well as portrayals of battles, the proclamation of the Rio Grandense Republic, a milestone in the revolution, and the importance of the female figure during the conflict. Regarding customs, we depict the ``Chimarrão", a typical beverage of the southern region, and, finally, the traditional attire of RS people.

With the concepts defined, images were collected to create a dataset. The quantity and diversity of these images depend on each of the concepts. For example, there is a large number of images regarding the biomes, 
the Araucaria trees, but a limited number of images regarding the Farroupilha Revolution and its related concepts, like the person of 
Anita Garibaldi.

A three-stage curatorship process was undertaken to generate the datasets intended for training. In the initial stage, a web-crawling algorithm was employed to gather images associated with the specified concepts. The second stage involved visual curation to ensuring in an initial set of images with high-quality. We established exclusion criteria to compose the training data, excluding i) images with low correlation to the given concept and ii) images with low DPI and resolution. The first criterion was necessary to assess the suitability of the samples for each concept. Table \ref{tab:training-images} presents the final number of images in each dataset.

\begin{table}[h]
    \centering
    \begin{tabular}{@{}cc@{}}
        \toprule
        \textbf{Concept} & \textbf{Trained Images} \\
        \midrule
        Cuia e Bomba de Chimarrão & 28 \\
        Indumentária Gaúcha & 84 \\
        Araucárias & 97 \\
        Pampas & 46 \\
        Gato-do-mato-pequeno & 41 \\
        Sapinho-admirável-de-barriga-vermelha& 31 \\
        Anita Garibaldi & 14 \\
        Bento Gonçalves & 12 \\
        Giuseppe Garibaldi & 22 \\
        Farroupilha Battles & 49 \\
        \bottomrule
    \end{tabular}
    \caption{Number of training images used for fine-tuning each concept}
    \label{tab:training-images}
\end{table}



\subsection{Training Details}
\label{training}

In the process of training our models, we implement the Dreambooth training script, available at Diffusers \cite{von-platen-etal-2022-diffusers}. It is relevant to highlight that we finetuned each dataset separately, generating a model that learned and specialized in that concept. For each dataset, we assigned a distinct token, creating a unique training instance prompt, such as "A painting of \textbf{[V]}". We used Stable Diffusion \cite{rombach2022high} v-1.4 and v-1.5 as a trained model.


The models were trained on the default resolution of $512 \times 512$ of SD v-1.4 and v-1.5, with a batch size of 1, using the AdamW optimizer. We experimented with different learning rate values ranging from $3 \times 10^{-6}$ to $5 \times 10^{-7}$. We also employed a ``constant" learning rate scheduler, ensuring a consistent rate throughout training. The number of training steps to achieve good results varies according to the dataset. In our experiments, we performed extensive testing within a range of $1.000$ to $10.000$ training steps. Fewer steps did not effectively capture the desired concepts, even though a substantial increase in the step count resulted in a model that became overly biased for specific concepts, limiting its ability to absorb additional prompts effectively. During inference, we used the PNDMScheduler\footnote{https://huggingface.co/docs/diffusers/api/schedulers/pndm} with 50 steps, 7.5 classifier-free guidance, at resolution $512 \times 512$.

All the fine-tuning experiments were conducted on instances using a single NVIDIA A100 GPU with 80GB of memory. Inference was carried out on a V100 GPU instance with 16 GB of memory.

\section{Results}
\label{results}
This section discusses the images obtained from the DreamBooth fine-tuning, dividing them into three results. First, focus on cultural aspects that explore the representation of the ``Cuia e Bomba de Chimarrão", essential elements of the mate tea culture, and the ``Indumentária Gaúcha" (traditional gaucho costume). Additionally, it explored the representation of fauna and flora, with emphasis on the ``Pampas" landscape, ``Araucárias" trees, and endangered endemic animals, such as the red-bellied frog (``Sapinho-admirável-de-barriga-vermelha") and the ``Gato-do-mato-pequeno" cat. The narrative of the Farroupilha Revolution delves deeper into historical figures. We also delve into some significant battles during this period. Our journey through these visual narratives showcases Dreambooth's ability to portray elements of Brazil's cultural, ecological, and historical events.

\subsection{Cultural Aspects}

\begin{figure*}
\centering
\includegraphics[width=\textwidth]{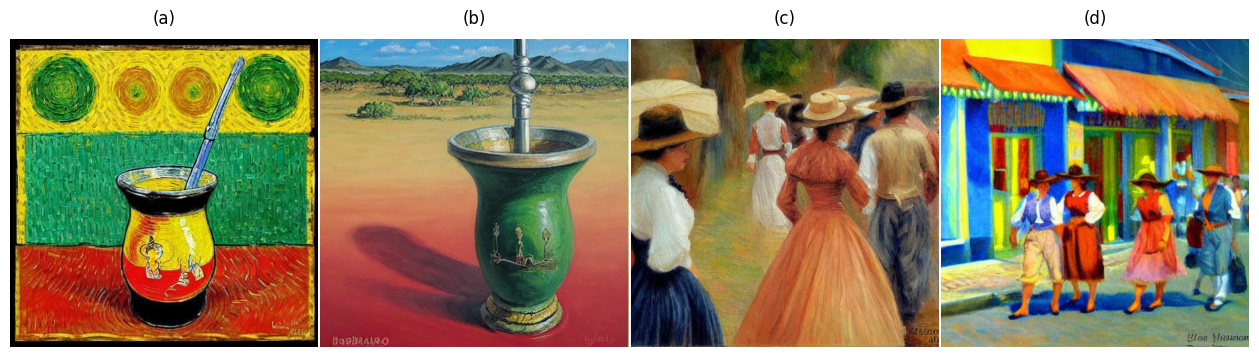}
\caption{Images generated by Dreambooth to represent the tea and clothing culture of Rio Grande do Sul, Brazil. The prompts adopted were (a) A painting of \textit{cuiaBombaChimarrao}, Van Gogh Style. (b) A painting of \textit{cuiaBombaChimarrao}. (c) A painting of \textit{indumentariaGaucha}, impressionism style. (d) A painting of \textit{indumentariaGaucha}, impressionism style.}
\label{fig1}
\end{figure*}

In the scope of cultural aspects, we have successfully derived subject-driven models capable of representing images of some of the customs of Rio Grande do Sul. As training concepts, we choose to explore the traditional drink of the southern region of Brazil, ``chimarrão" and the traditional attire of Gaúchos. The chimarrão is a mate-based drink introduced by the Guarani indigenous people, making it a deep-rooted typical drink. The ``cuia" and ``bomba de chimarrão", usually known as "mate straw" \cite{bernardes2021chimarrao}, are used for the preparation and consumption of the drink. Figure \ref{fig1} (a) portrays a ``Cuia de Chimarrão" in the style of Van Gogh, with the colors of the Rio Grande do Sul flag in the background, while Figure \ref{fig1} (b) portrays a painting without a specified artistic style, demonstrating Dreambooth's versatility in creating diverse visual representations. Regarding the latter concept, the traditional attire, or ``indumentária gaúcha" is characterized by its unique elements, which include ``bombacha" (Refers to the loose-fitting, baggy trousers or pants, often made of durable fabric), ``gaita" (sash or belt), a shirt, traditional leather boots, a neckerchief or bandana, wide-brimmed hat (often known as the ``chimarrão" hat), ``poncho" (a shoulder-worn fabric piece), and a waistband or sash. This clothing encompasses the traditional man's attire. On the other hand, the ``prenda" symbolizes the representation of the rural woman, including long skirts, ornate blouses, scarves, and boots. These components collectively make up the traditional attire of gaúchos, serving both functional and cultural purposes. Figures \ref{fig1} (c) and \ref{fig1} (d) depict some of these elements of gaúcho attire in the impressionist style.

Regarding the ``chimarrão" a substantial number of images containing a ``cuia" and ``bomba de chimarrão" were found to build the database. However, a significant challenge arose due to several featured logos, resulting in distorted logos in the output domain of the model. Consequently, the curatorship process had to filter out items without logos for the database. Another issue pertained to the close-up shots of many photos depicting the concept, with the object occupying a substantial portion of the image. This proximity caused the model to lack a realistic sense of the mate gourd's size. When asked the model to depict a person drinking mate, the mate gourd would appear larger than its actual size, or the model failed to understand that the person would drink through a straw, treating the mate gourd as if it were a cup. Furthermore, due to the similar shape, the model occasionally added flowers inside the mate gourd, as if it were a plant pot. After extensive experimentation with different configurations, we generated highly satisfactory images related to the concept.

In the case of ``indumentária gaúcha" (traditional attire), there was also an abundance of images available on the internet. However, a significant challenge emerged as the clothing from various periods often became intertwined, blending with more contemporary adapted pieces. Additionally, certain concepts like "bombacha" and "gaita" often had to be included in separate images (not worn by individuals but displayed in showcases, for example) to enable the model to capture their specific details. Moreover, the prevalence of images featuring multiple individuals wearing the traditional attire, often from events where dozens of people wear these garments, seemed to bias the model towards generating images of groups of people in the attire rather than focusing on one or two individuals. Despite these challenges, the model captures the essence of the attire remarkably.

\subsection{Fauna and Flora}

\begin{figure*}
\centering
\includegraphics[width=\textwidth]{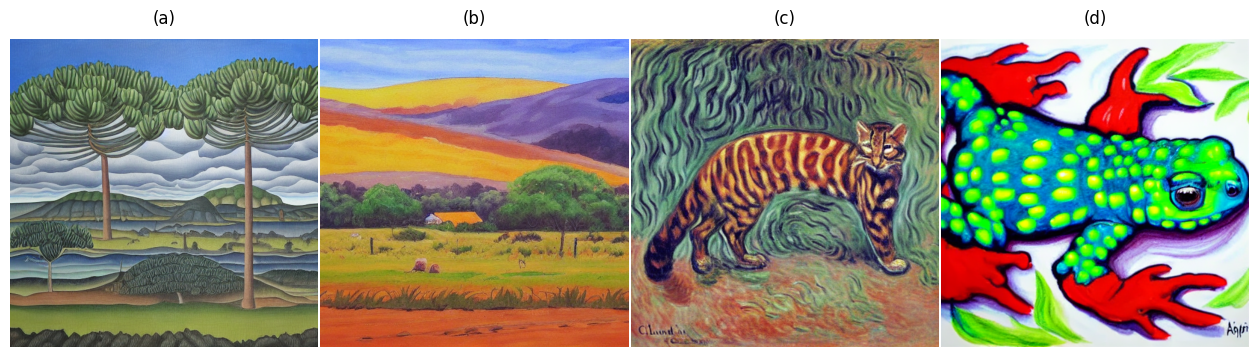}
\caption{Images generated by Dreambooth to represent the fauna and flora of Rio Grande do Sul, Brazil. The prompts adopted were: (a) A painting of \textit{araucaria}, Tarsila do Amaral Style. (b) A painting of \textit{Pampa} landscape. (c) A painting of \textit{gatoDoMato}, Claude Monet Style. (d) A painting of \textit{sapinhoBarrigaVermelha}, kid's drawing style.}
\label{fig2}
\end{figure*}

In our exploration of fauna and flora, we investigated the potential use of AI to generate images of endangered species, specifically the red-bellied toad (\textit{Melanophryniscus admirabilis} or ``Sapinho-admirável-de-barriga-vermelha") and the little spotted cat (\textit{Leopardus tigrinus} or ``Gato-do-mato-pequeno"). The first is an endemic species with its habitat limited to a 700-meter stretch along the Forqueta River in the Arvorezinha municipality of RS \cite{SapinhoAdmiravel}. The latter originates from Central and South America and is endangered due to deforestation and habitat conversion to agricultural land \cite{Silveira2018}. Figure \ref{fig2} (c) depicts the ``Gato-do-mato-pequeno" in Claude Monet style, whereas Figure \ref{fig2} (d) represents the ``Sapinho-admirável-de-barriga-vermelha" rendered in the style of a child's drawing. For the regional flora, we chose to represent ``Araucárias", a group of evergreen trees native to South America and part of the Atlantic Forest, which has experienced severe deforestation, with only $1-4\%$ of its original area remaining \cite{mantovani2004fenologia}. Additionally, we focused on the ``Pampas", extensive natural plains covering $2\%$ of Brazilian territory. Figures \ref{fig2} (a) depict ``Araucárias" in the style of Tarsila do Amaral's paintings, and Figure \ref{fig2} (b) shows a painting of ``Pampas" landscapes.

Regarding the ``Sapinho-admirável-de-barriga-vermelha", several challenges were encountered. Images often portrayed the toad in a close perspective, causing the model to lose its sense of scale, similar to the issue with the mate gourd. The model tended to confuse the color of the toad's belly/legs with the ground, often depicted as red. Furthermore, there were difficulties in accurately representing the toad's limbs and toes. This concept proved to be particularly challenging to illustrate. The main challenges for the ``Gato-do-mato-pequeno" concept revolved around its similarity to other animals already known to the model, such as domestic cats, ocelots, and others. To address this, we ensured the model had sufficient images of the distinctive features of this species in the training dataset and depicted them in their natural habitat.

In the case of ``Araucárias", the main challenge lay in appropriately curating the dataset to ensure that other tree species were not present in the data, which could potentially bias the model. Many photos depicted a single tree in the center of the image, which tended to skew the model's preferences towards this configuration, resulting in less successful outcomes when the representation involved denser vegetation. Overall, this was one of the concepts where the model presented an easy learning process and demonstrated greater flexibility in our study. For the ``Pampas", the primary challenge arises from the vast variability of images. Nevertheless, the model grasped that the concept pertained to a landscape and its fundamental characteristics.

\subsection{Farroupilha Revolution}

This section delves into the Farroupilha Revolution period in Brazilian history, especially for Rio Grande do Sul. Through images generated by Dreambooth, we immerse ourselves in the lives and actions of the characters who shaped this tumultuous period. In the first subsection, we present generated images that capture pivotal moments in the lives of Anita Garibaldi, Giuseppe Garibaldi, and Bento Gonçalves. The second subsection focuses on the battles that marked the Farroupilha Revolution. We explore everything from the proclamation of the Rio Grandense Republic to the Crossing of the Lanchões, highlighting the details of these historical events. Additionally, we underscore the often underestimated role of women in the revolution, demonstrating their vital contribution to the movement.

\subsubsection{Historical Characters}
\begin{figure*}
\centering
\includegraphics[width=\textwidth]{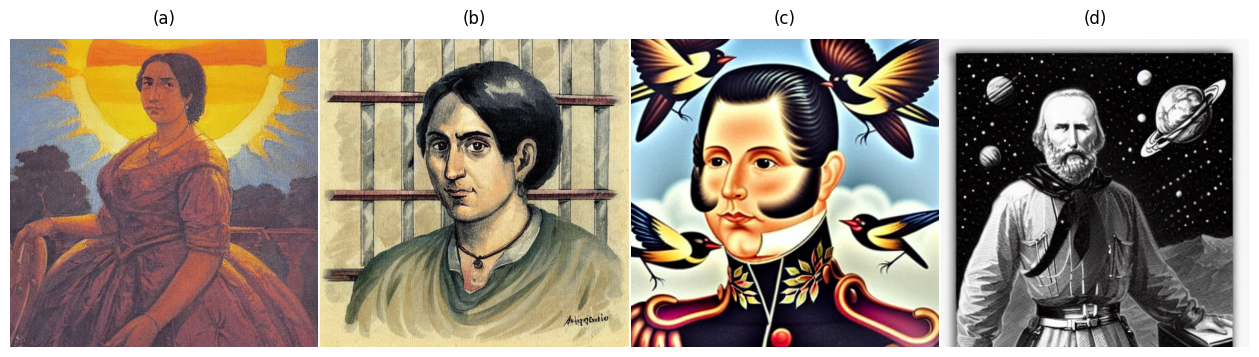}
\caption{Dreambooth generated images to represent the Farroupilha revolution. The prompts used were: (a) A painting of \textit{aniGaribaldi}, with a big sun behind her, (b) A painting of \textit{aniGaribaldi}, kept in prison, (c) A painting of \textit{bentoGonçalves}, with birds around him. (d) A painting of \textit{giuseppeGaribaldi}, in outer space, with planets and stars around him.}
\label{fig3}
\end{figure*}

The Farroupilha Revolution, which unfolded between 1835 and 1845 in the southern region of Brazil, centered on Rio Grande do Sul State. It was a movement that sought substantial reforms, including the quest for autonomy to reflect the region's unique characteristics and demands \cite{zalla2011historia}. Led by iconic figures such as Anita Garibaldi, Bento Gonçalves, and Giuseppe Garibaldi, this episode transcended the boundaries of southern Brazil, leaving an indelible mark on the country's history. Anita Garibaldi is depicted in Figure \ref{fig3} (a) with a sunset as a backdrop, Figure \ref{fig3} (b)  captured by opposing forces. Bento Gonçalves, depicted in Figure \ref{fig3} (c), was one of the leading military leaders of the Farroupilha Revolution. His role was instrumental in shaping the course of the rebellion, contributing significantly to the pursuit of autonomy for the southern Brazilian region. Giuseppe Garibaldi, represented in Figure \ref{fig3} (d), also known as "The Hero of Two Worlds" due to his participation in the unification of Italy and central role in Rio Grandense independence, is an influential character for the southern Brazilian people \cite{oliveira2012garibaldi}. Alongside Bento Gonçalves, he played a crucial role in leading the revolution and advocating for the unique characteristics and demands of the region. 

Images of historical figures from the 19th century, such as Anita Garibaldi, Giuseppe Garibaldi, and Bento Gonçalves, are notably scarce in terms of variety. The existing photographs often provide limited perspectives, with some representations focusing only on the bust upwards. Furthermore, many historical images lack sufficient DPI and resolution, presenting challenges in capturing the filled essence of these figures. The limited visual records of these personalities create a scarcity of resources, particularly when considering the diverse and comprehensive representations. Despite these challenges, the model development centered on these historical subjects opens up new possibilities. These models, incorporating rare tokens like \textit{aniGaribaldi}, \textit{GiusepGaribaldi}, and \textit{bentoGoncalves} within instance training prompts, demonstrate a remarkable ability to generate artistic images. 

Figure \ref{fig3} showcases the model's capacity to derive visually pleasing and conceptually rich images, even in scenarios with sparse historical representations. With fine-tuning configurations and inferences, the model navigates adversities posed by limited visual resources. Enable the creation of artistic depictions but also allow the integration of various elements, such as animals, into the narrative of the image. In essence, the ability to represent historical figures with scarce visual resources underscores the power of AI models in preserving and promoting historical narratives. By leveraging the potential of these models, we can bridge gaps in visual records, offering contemporary audiences a more holistic and engaging insight into the vital roles played by individuals such as those. This technological advancement contributes significantly to preserving and celebrating the cultural heritage of nations.

\subsubsection{Battles}

The Farroupilha Revolution, a defining period in Brazilian history, was marked by numerous battles that played a pivotal role in the struggle for independence. These battles were the crucible in which the aspirations of the rebels clashed with the forces of the Brazilian Empire. They decided the fate of the revolution but also contributed to the shaping of Brazil's future \cite{sinotti2015revoluccao}. Some of the crucial events of the Farroupilha Revolution were the "Travessia dos Lanchões" or Crossing of the Lanchões, vessels driven by land, on wheels and pulled by cattle, and the Proclamation of the Rio Grandense Republic, a nation-state separate from the Brazilian state.

\begin{figure*}
\centering
\includegraphics[width=\textwidth]{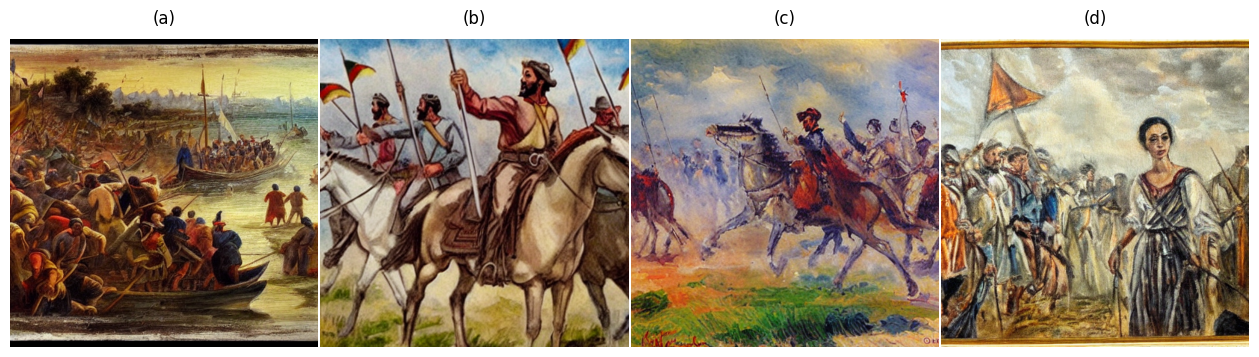}
\caption{Dreambooth generated images to represent historical battles from the Farroupilha revolution. The prompts used were: (a) A painting of \textit{batalhaFarroupilha}, crossing boats. (b) A painting of \textit{batalhaFarroupilha}, republic ploclamation. (c) A painting of \textit{batalhaFarroupilha}, republic ploclamation. (d) A painting of \textit{batalhaFarroupilha}, a woman in the center of the battle.}
\label{fig4}
\end{figure*}

Figure \ref{fig4} (a) captures the daring Crossing of the "Lanchões" under the leadership of Giuseppe Garibaldi, hauled the boats, or "lanchões," several kilometers over land, through dense terrain and across rugged landscapes, reaching the destination and crossing the Guaíba River, thereby outmaneuvering the imperial forces. Images serve as a visual testament to the audacity and resourcefulness that marked this episode in the Farroupilha Revolution. Figures \ref{fig4} (b) and \ref{fig4} (c) depict the iconic moment of the "Proclamation of the Republic" during the Revolution. It was a declaration that echoed the rebels' commitment to the cause of independence, asserting their determination to create a republican government. This act was a defining moment in the Revolution's narrative. The "Proclamation of the Republic" was a crucial event in the Farroupilha Revolution's historical context. It took place on September 11, 1836, when the revolutionary forces, led by Bento Gonçalves, declared the independence of Rio Grande do Sul from the Brazilian Empire and the establishment of a republic.  Figure \ref{fig4} (d) pays tribute to an often-overlooked aspect of the Farroupilha Revolution: the significant role of women in the conflict. Women played diverse and crucial roles during this period, including serving as dedicated nurses and providing essential medical care to wounded soldiers. They were instrumental in managing logistics and procuring and distributing vital supplies. Some acted as couriers, relaying messages and contributing to intelligence efforts. Additionally, women supported troops by maintaining camps, preparing meals, and ensuring soldiers' well-being. They also undertook the solemn task of burying fallen soldiers, ensuring a proper and respectful farewell.

Much of the visual documentation of these occurrences comes from paintings. Typically, the scenes depicted in these images consist of numerous mounted combatants, lacking a central subject in the visual samples. Similarly to the historical personality datasets, these concepts are scarce in terms of the quantity, diversity, and quality of images; however, DreamBooth generates satisfactory images. For battles generation, we trained our models with the Batalhas da Farroupilha dataset that includes images of generic battles (most of them), the proclamation of Rio Grandense republic, Crossing of the Lanchões, and also, images of the feminine representation of the Revolution.





\section{Conclusion}
\label{conclusion}
In conclusion, our comprehensive exploration of Dreambooth's generative capabilities, framed within the context of Rio Grande do Sul has yielded promising outcomes. The fine-tuning process allowed us to capture the essence of Brazilian culture, showcasing elements such as ``cuia e bomba de chimarrão", the ``indumentária gaúcha", the lush fauna and flora, and the rich historical tapestry of the Farroupilha Revolution. The generated images, with technically proficient and artistically captivating, underscore Dreambooth's remarkable potential in bridging the realm of text-based prompts and visual art.

Throughout our experimentation, specific areas for improvement and best practices have come to light:

\begin{enumerate}
    \item \textbf{Exploration of Diverse Training Configurations}: We recommend delving deeper into a broad range of training configurations to identify superior settings for image generation. Additionally, further exploration of hyperparameters can help enhance the overall quality of the generated images.  \item \textbf{Diversification of Prompts}: By refining and expanding the input prompts, one can produce a more varied and evocative array of images, further enriching the user experience.
\end{enumerate}

In conclusion, it is important to emphasize that the elements presented here are not inherently valuable as standalone works, but they serve as illustrative examples of how generative AI can be harnessed to represent crucial concepts for various communities. For instance, the chosen endemic animals are currently facing extinction, and the ability to generate representations from the available data in a repository could hold significant long-term importance. Similarly, the historical figures and battles depicted here have long faded into the past, with very few accessible images in public archives. In this context, generative AI can play a vital role in creating more visual content related to such events and individuals, for educational purposes or other potential applications. Furthermore, we must acknowledge the potential for blending various historical art movements' styles, such as expressionism and modernism, or even the techniques of long-deceased artists like Van Gogh and Tarsila do Amaral. In essence, these examples underscore how diffusion models can be employed to generate representative art for local communities, allowing for the preservation and revitalization of cultural and historical significance.

\section*{Acknowledgments}
We acknowledge the support from Conselho Nacional de Desenvolvimento Científico e Tecnológico (CNPq), and 
Alana AI for funding the research. We would also like to express our heartfelt appreciation to our team of linguists who curated the dataset used in the training of our models.

\section*{Funding}

This project was funding by Conselho Nacional de Desenvolvimento Científico e Tecnológico (CNPq) and Alana AI. 


\section*{Contributions}
AZ and FM conducted the experiments. WC and MS provided oversight for the research. All authors contributed to the paper's composition, with MS and WC contributing to the study's conception. The final manuscript was reviewed and approved by all authors.

\section*{Interests}
The authors declare that they have don't have any competing interests.

\section*{Materials}
The datasets generated and/or analysed during the current study are available by request.

\bibliographystyle{apalike-sol}
\bibliography{main}

\end{document}